\pdfoutput=1

\documentclass[11pt]{article}

\usepackage[final]{acl}

\usepackage{times}
\usepackage{latexsym}
\usepackage{hyperref}
\usepackage{todonotes}
\usepackage{enumitem}
\usepackage{tabularx}
\usepackage{amsmath} 
\usepackage{booktabs}

\usepackage[T1]{fontenc}

\usepackage[utf8]{inputenc}

\usepackage{microtype}

\usepackage{inconsolata}

\usepackage{graphicx}

\usepackage{listings}
\usepackage{xcolor}
\usepackage{float}
\title{Soft Measures for Extracting Causal Collective Intelligence}

\author{
 \textbf{Maryam Berijanian\textsuperscript{1}},
 \textbf{Spencer Dork \textsuperscript{1,3}},
 \textbf{Kuldeep Singh\textsuperscript{2}},
 \textbf{Michael Riley Millikan\textsuperscript{1}},
\\
 \textbf{Ashlin Riggs\textsuperscript{1}},
 \textbf{Aadarsh Swaminathan\textsuperscript{1}},
 \textbf{Sarah L. Gibbs\textsuperscript{4}},
 \textbf{Scott E. Friedman \textsuperscript{5}},
\\
 \textbf{Nathan Brugnone\textsuperscript{1,3}}
\\
\\
 \textsuperscript{1}Department of Computational Mathematics, Science, \& Engineering, Michigan State University
 \\
 \textsuperscript{2}Department of Computer Science, Michigan State University
 \\
 \textsuperscript{3}Complex \& Social Systems Lab, Two Six Technologies,
 \textsuperscript{4}University of South Alabama,
 \textsuperscript{5}SIFT
\\
 \small{
   \texttt{\{berijani, dork, singhku2, millika6, riggsash, swamina9, brugnone\}@msu.edu}
 }
 \\
 \small{\texttt{slg2221@jagmail.southalabama.edu}, \texttt{friedman@sift.net}}
}

\begin{document}
\maketitle
\begin{abstract}

Understanding and modeling collective intelligence is essential for addressing complex social systems. Directed graphs called fuzzy cognitive maps (FCMs) offer a powerful tool for encoding causal mental models, but extracting high-integrity FCMs from text is challenging. This study presents an approach using large language models (LLMs) to automate FCM extraction. We introduce novel graph-based similarity measures and evaluate them by correlating their outputs with human judgments through the Elo rating system. Results show positive correlations with human evaluations, but even the best-performing measure exhibits limitations in capturing FCM nuances. Fine-tuning LLMs improves performance, but existing measures still fall short. This study highlights the need for soft similarity measures tailored to FCM extraction, advancing collective intelligence modeling with NLP.

\end{abstract}



\section{Introduction}


Social science has long sought to understand and model the collective intelligence underlying humanity's most pressing problems such as climate change, sustainable food supply, and violent conflict driven by inequitable resource distribution.
These are \textit{social-ecological systems} (SES) problems characterized by complex, interwoven feedback loops involving human and natural systems \cite{ostrom2009general, partelow2018review}.
To model collective intelligence about SES, we can leverage mental models of causal system structure.

\begin{figure*}[h]
    \centering
    \includegraphics[width=1.0\linewidth]{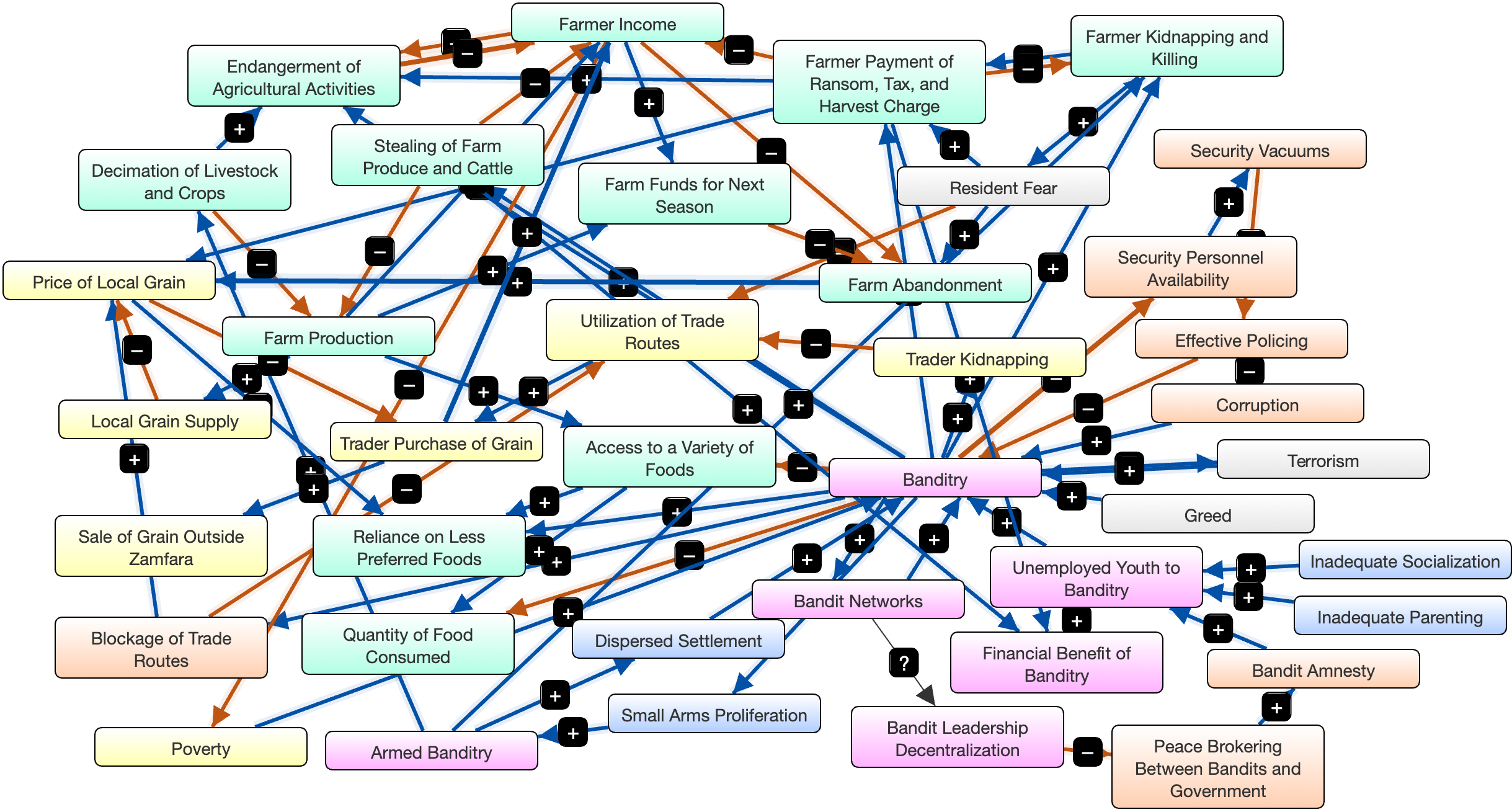}
    \caption{FCM describing the \citet{mmahi2023brigandage} mental model of conflict and food system dynamics in Zamfara State, Nigeria; blue (resp. red) edges indicate causally increasing (resp. decreasing) relations; visualized with \href{https://www.mentalmodeler.com/}{Mental Modeler}.}
    \label{fig:fcm_example}
\end{figure*}

Researchers in the social sciences have formally encoded SES mental models using \textit{fuzzy cognitive maps} (FCMs) that represent causal systems as signed, weighted digraphs, where edges represent causal relationships among natural language concepts \cite{kosko1986fuzzy} like that depicted in Figure \ref{fig:fcm_example}.
FCMs are inspired by human \emph{causal mental models} that people use to explain causal mechanisms and generate predictions \cite{craik1967nature}.


FCMs have been widely used to facilitate cross-disciplinary communication within research teams \cite{gray2013fuzzy}, make qualitative and numerical predictions, and assess collective intelligence \cite{gray2020harnessing, aminpour2020wisdom, Voinov2018}.
However, little work has explored extracting high-integrity FCMs from textual corpora. The development of a robust text-to-FCM method would enable more rapid synthesis of science- and stakeholder-informed perspectives to provide access to latent collective intelligence about SES.

This work presents a \textit{natural language processing} (NLP) approach to (1) extracting FCMs from text with \textit{large language models} (LLMs) and (2) measuring extracted FCM quality with novel \emph{soft} F1 measures that permit approximate semantic matches rather than requiring exact node and edge matches.
This helps capture and accumulate diverse causal collective intelligence of SES domains.
We provide code and datasets for reproducibility.\footnote{The source codes and dataset are available at \href{https://github.com/kuldeep7688/soft-measures-causal-intelligence}{https://github.com/kuldeep7688/soft-measures-causal-intelligence}.}

\section{Background and Objectives}
Extracting FCM edges is a specialization of extracting semantic relations or causal graphs: each node is a textual span describing causal factors, and each edge is a directed causal \emph{increase} or \emph{decrease} relationship where the source of the edge quantitatively increases or decreases the target. Two issues distinguish FCM extraction from previous graph-based extraction tasks: (1) high expressiveness and (2) partial correctness.  We describe these two issues briefly to motivate our approach.

\paragraph{High expressiveness.} Causal variables (i.e., FCM nodes) and relations may be expressed multiple ways, so human annotators may disagree and a machine prediction may actually outperform a pre-determined human-generated \textit{gold standard} according to human judges.
Many graph extraction approaches are validated by their proximity to a singular gold standard generated by human experts, where small deviations from the standard greatly affect the measure. This includes neural network loss functions. Some relaxed matching strategies help account for textual containment or overlap \cite{Chen2019, Toba2010}, but these have not been adequately extended to graph extraction.

\paragraph{Partial correctness.}  A sub-optimal node or edge that captures a valid causal relationship is often useful to report as a component of an FCM---especially in data-poor contexts, as even limited information can improve the understanding of a given system---so an NLP model's capability to produce partially correct edges is important to capture (Table \ref{table:causal_relationships}).

\begin{table}[h]
\centering
\footnotesize
\begin{tabular}{p{1.1cm} p{1.3cm} p{0.6cm} >{\raggedright\arraybackslash}p{2.5cm} } 
\toprule 
\multicolumn{1}{l}{\textbf{Source}} & \multicolumn{1}{l}{\textbf{Target}} & \multicolumn{1}{l}{\textbf{Direction}} & \multicolumn{1}{l}{\textbf{Issue}} \\ 
\midrule 
turbine \newline structures & blue \newline mussels & increase & NA (gold standard)\\  
\midrule  
numbers\newline of blue \newline mussels & turbine \newline structures & increase & source \& target swapped; extra source text \\ 
\midrule  
turbines & mussel \newline populations & decrease & simplified source; extra target text; incorrect direction \\ 
\midrule
turbine \newline structures & blue \newline mussels & decrease & incorrect direction \\ 
\bottomrule 
\end{tabular}
\caption{Examples of partially correct causal relationships. Original text: "Some fishermen described the establishment of large numbers of blue mussels on the turbine structures" \cite[p. 245]{ten2021integrating}.}
\label{table:causal_relationships}
\end{table}

Consequently, binary judgments of correctness---such as precision, recall, and F1 scores---are sub-optimal measures for our task.  Decades of research has produced numerical measures to score the similarity (or distance) between spans of text to avoid the need for absolute correctness \cite{Mihalcea2006, Bar2012, Lavie2009}, but development of textual similarity measures for graphs, and FCMs in particular, has been limited \cite{Pilehvar2015}.

This paper (1) assesses fine-tuned LLM-based methods to extract FCMs from text and (2) introduces and evaluates edge-based similarity measures for validating FCM quality, addressing the limitations of previous measures. Additionally, it (3) initiates an approach for validating graph-based NLP predictions by (a) ranking predictions through pairwise comparative human judgments using Elo and (b) comparing the rankings produced by humans and similarity measures. This proof-of-concept study suggests a methodology by which to improve the qualitative evaluation of NLP-generated FCMs en masse and, thereby, takes a step towards improved collective intelligence models.

\section{Methods and Data}
\subsection{Dataset}
We curated a dataset of 318 short text passages extracted from a diverse set of research articles on SES. These articles cover a wide range of SES topics including offshore wind farm development, the impact of banditry on the food system in northern Nigeria, the distribution of food and medical aid in conflict regions, and maternal and child health in countries with low Human Development Index scores. We annotated each text passage with (\texttt{source, target, direction}) tuples.


\subsection{Annotation Ranking}
To rank annotations for each text passage, we (1) generated multiple annotations for each of a subset of passages, (2) presented pairs of annotations to raters, and (3) applied the Elo rating system.
\subsubsection{Annotation Generation} 
Each of a subset of 20 passages were manually annotated with (\texttt{source, target, direction}) tuples by all authors. This subset was further augmented with LLM annotations. This was achieved through two distinct methods: few-shot learning and instruction tuning with LoRA \cite{instruction, Hu2021}. We employed the Llama-2-7B-chat-hf \cite{LLaMAseries, llama2}, Llama-3-8B-Instruct \cite{llama3}, and Mistral-7B-Instruct-v0.2 \cite{mistral} models from Hugging Face \cite{huggingface}. Fine-tuning was accomplished using splits of the 318 data points. A detailed presentation of these methods appears in Appendices \ref{zero_three_shot_prompt}, \ref{parameters}, and \ref{prompt}.

\subsubsection{The Elo Rating System}
The Elo rating system, introduced by Arpad Elo \cite{elo, eloBook}, is a widely used method for quantifying the relative skill levels of players in two-player competitive games. 
It has been demonstrated to effectively rank models based on human judgment \cite{eloUncovered}, benchmark LLMs \cite{chatbot}, and rank preferences, such as humor in Twitter posts \cite{chatbot} through pairwise comparisons. Inspired by previous research \cite{fitnessFunctionPaper, whichIsBetter}, we used the Elo system to rank annotations and then compared these rankings with those generated by the candidate similarity measures.



\subsubsection{Elo Tournaments}
Each author was presented with a series of comparisons between annotations via a web interface (Appendix \ref{app:UI}). Raters were instructed to select the better annotation as `winner' or to choose `tie' following a set of guidelines (see Appendix \ref{app:rater-guidelines}). Elo scores were computed \textit{per passage}, so each passage acted as an individual `tournament'. Raters did not rate their own annotations to avoid potential bias. Inter- and intra-rater reliability were captured through overlaps (Appendix \ref{app:inter-rater}). 

\subsection{Similarity Measures}
We devised five candidate FCM similarity measures based on the established textual similarity measures in Table \ref{table:kg_metrics}. The similarity between an FCM and a gold standard is computed as a \textit{softly thresholded F1 score} between edge sets. Given a textual similarity measure $S\left(\cdot,\cdot\right)$, a threshold $T$, and edge sets $E$ and $E_{\text{gold}}$, as well as any textual edge attributes $\texttt{A}$ (we use $\texttt{A} = \{\texttt{source}, \texttt{target}\}$) and non-textual edge attributes $\texttt{N}$ (we use $\texttt{N} = \{ \texttt{direction}\}$):
\begin{enumerate}
    \item[\textbf{TP:}] For each $e\in E$, our method counts a \textit{true positive} if there exists an $e_{\text{gold}}\in E_{\text{gold}}$ such that $S\left(e.\texttt{a},e_{\text{gold}}.\texttt{a}\right)\geq T$ for every $\texttt{a}\in\texttt{A}$ and $e.\texttt{n}=e_{\text{gold}}.\texttt{n}$ for every $\texttt{n}\in\texttt{N}$ ;
    \item[\textbf{PP:}] For each $e\in E$, our method counts a \textit{partial positive} if there exists an $e_{\text{gold}}\in E_{\text{gold}}$ such that $S\left(e.\texttt{a},e_{\text{gold}}.\texttt{a}\right)\geq T$ for every $\texttt{a}\in\texttt{A}$ and there exists 
 an $\texttt{n}\in\texttt{N}$ such that $e.\texttt{n}\neq e_{\text{gold}}.\texttt{n}$;
    \item[\textbf{FP:}] For each $e\in E$, our method counts a \textit{false positive} if for every $e_{\text{gold}}\in E_{\text{gold}}$, we have $S\left(e.\texttt{a},e_{\text{gold}}.\texttt{a}\right)<T$ for any $\texttt{a}\in\texttt{A}$;
    \item[\textbf{FN:}] For each $e_{\text{gold}}\in E_{\text{gold}}$, our method counts a \textit{false negative} if for every $e\in E$, we have $S\left(e.\texttt{a},e_{\text{gold}}.\texttt{a}\right)<T$ for any $\texttt{a}\in\texttt{A}$.
\end{enumerate}
For any $S$, once the TP, PP, FP, and FN have been counted, the corresponding edge-based measure can be calculated using the F1-like formula:
\begin{equation}
\frac{2 \cdot \text{TP} + \text{PP}}{2 \cdot \text{TP} + \text{PP} +\text{FP} + \text{FN}}.
\label{eq:f1}
\end{equation}
For each $S$, a threshold $T$ is chosen by grid search (Appendix \ref{app:similarity-hyperparams}). Note that by disallowing for partial positives, when $S\left(\cdot,e.\texttt{a}\right)=\mathbf{1}_{e.\texttt{a}}\left(\cdot\right)$---an exact match criterion---and $T=1$, this reduces to the classical F1 score between edge sets.

For BLEU, ROUGE, METEOR, and BLEURT, we refer to our novel edge-based measures as BLEU-E, ROUGE-E, METEOR-E, and BLEURT-E, respectively.

\begin{table}[tb]
\centering
\footnotesize
\begin{tabular}{p{1.6cm} p{2.2cm} p{2.6cm}}
\toprule
\multicolumn{1}{l}{\textbf{Measure}} & \multicolumn{1}{l}{\textbf{Strengths}} & \multicolumn{1}{l}{\textbf{Weaknesses}} \\ \midrule
Exact match & Simple and direct measurement. & Sensitive to minor textual differences. \\ \midrule
BLEU \cite{papineni-etal-2002-bleu} & Considers n-gram precision and brevity. & Does not account for synonyms or grammatical meaning. \\ \midrule
ROUGE \cite{lin-2004-rouge} & Flexible in measure type and n-gram method. & Does not account for synonyms or grammatical meaning. \\ \midrule
METEOR \cite{banerjee-lavie-2005-meteor} & Accounts for synonyms, stems, and word order. & Does not account for context or grammatical meaning. \\ \midrule
BLEURT \cite{sellam2020bleurt} & Captures abstract meanings using neural networks. & Potential biases and limited user control. \\ \midrule
\end{tabular}
\caption{Text similarity/matching strategies.}
\label{table:kg_metrics}
\end{table}

\subsection{Correlation Analysis}

The winning annotation of each tournament was deemed the gold standard. We then produced a ranking of annotations \textit{per passage} using each candidate similarity measure applied to each (\texttt{gold standard, annotation}) FCM pair. The Spearman correlations \cite{spearman} between human- and similarity measure-generated rankings were computed. We then applied the measure with highest correlation to evaluate LLM-generated FCMs and compared with an LLM-only tournament.

\section{Results}
\subsection{Spearman Correlation}

The Spearman correlation coefficients for each measure, averaged across all passages, are summarized in Table \ref{table:spearman_results}. Higher values indicate greater mean correlation with human rankings. Novel measure-produced rankings have positive mean correlations with human-generated rankings, and each improves upon vanilla F1 in this regard (Table \ref{table:spearman_diff_results}). 

\begin{table}[h]
\centering
\footnotesize
\begin{tabular}{l l l l}
\toprule
\textbf{Measure} & \textbf{Mean} & \textbf{90\% CI} & \textbf{95\% CI}\\ \toprule
F1 & 0.016 & (-0.057, 0.089) & (-0.072, 0.104) \\ 
BLEU-E$^*$& 0.109 & (-0.018, 0.237) &  (-0.045, 0.263) \\ 
METEOR-E$^*$ & \textbf{0.126} & (\textbf{0.001}, \textbf{0.252}) & (-0.025, 0.278) \\ 
ROUGE-E$^*$ & \textbf{0.124} & (\textbf{0.007}, \textbf{0.241}) & (-0.018, 0.266) \\ 
BLEURT-E$^*$ & \textbf{0.152} & (\textbf{0.038}, \textbf{0.265})  & (\textbf{0.014}, \textbf{0.289}) \\ 
\midrule
BLEU-E & \textbf{0.415} & (\textbf{0.257}, \textbf{0.574}) & (\textbf{0.223}, \textbf{0.607}) \\ 
METEOR-E & \textbf{0.333} & (\textbf{0.146}, \textbf{0.520}) & (\textbf{0.106}, \textbf{0.559}) \\ 
ROUGE-E & \textbf{0.387} & (\textbf{0.205}, \textbf{0.570}) & (\textbf{0.166}, \textbf{0.608}) \\ 
BLEURT-E & \textbf{0.338} & (\textbf{0.178}, \textbf{0.498}) & (\textbf{0.144}, \textbf{0.532}) \\
 \bottomrule
\end{tabular}
\caption{Mean correlations of similarity measures with human judgment and their confidence intervals. E$^*$ scores are computed without partial positives.}
\label{table:spearman_results}
\end{table}


\subsection{LLM Inferences}

\begin{figure}
    \centering
    \includegraphics[width=1\linewidth]{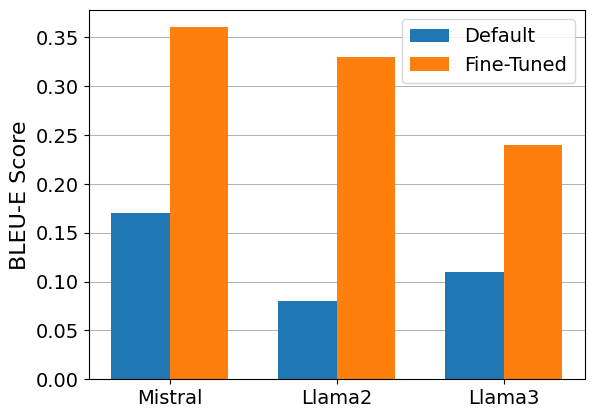}
    \caption{Mean BLEU-E across model variants.}
    \label{fig:e-bleurt-llms}
\end{figure}

Figure \ref{fig:e-bleurt-llms} presents the average BLEU-E scores for FCM inferences on the test set by each LLM before and after fine-tuning. As expected, fine-tuned models outperform their default counterparts, with Mistral scoring highest, followed by Llama-2 and then Llama-3.is consistent with the human-generated ranking. 

\section{Discussion and Conclusions}

This paper presents an evaluation of fine-tuned LLM-based methods for extracting FCMs from text, while also introducing and assessing novel edge-based similarity measures to validate the quality of these FCMs. The study highlights the limitations of traditional measures, which often fail to capture the nuances and partial correctness in FCMs, especially in the context of SES research. For instance, there were cases where LLMs outperformed human annotators, yet these instances might have been overlooked if we relied solely on traditional measures like the F1 score or validation set loss. The novel edge-based measures allowing for partial positives show markedly greater correlation with human judgments.


In addition to examining similarity measures, this work initiates a new approach for validating graph-based NLP predictions by employing pairwise comparative human judgments, using the Elo rating system, to rank predictions. This method was used to compare human-derived rankings with those generated by similarity measures, providing a proof of concept for enhancing the qualitative evaluation of NLP-generated FCMs.

Fine-tuning LLMs proved beneficial, resulting in higher BLEU-E scores and improved model performance. Whereas the validation set losses for all fine-tuned LLMs appear similar (Figure \ref{fig:ranks} in Appendix \ref{parameters}),  their qualitative performances differed significantly. Although BLEU-E offers a more accurate assessment than validation set loss, these improvements do not fully resolve the underlying issues with the current similarity measures. 

The study’s findings emphasize the necessity of developing more specialized measures that are better aligned with human judgment and capable of capturing the complexities of FCM extraction. This study represents an initial step towards that goal, highlighting the limitations of current approaches and setting the stage for future research.

Future work will focus on developing and validating new similarity measures that can better capture the complexities and partial correctness in FCM extraction. For instance, greater correlation with human judgment should be achievable by parameterizing TP and PP with scalars and employing optimization. Additionally, integrating human-in-the-loop approaches may help refine LLM outputs, leading to more accurate FCMs. We also imagine a range of applications and extensions. For instance, the proposed measures can be straightforwardly extended to knowledge hypergraphs. Furthermore, by symmetrizing any one of our typically asymmetric measures we may interpret it as a kernel \cite{kriege2020survey, scholkopf2018learning}. Explicitly, let $f=\text{BLEURT-E}$ and $G_1$ and $G_2$ be FCMs, and define a kernel $K$ as,
\begin{equation*}K\left(G_1,G_2\right)=\frac{f\left(G_1,G_2\right)+f\left(G_2,G_1\right)}{2}.
\end{equation*}
This interpretation brings to bear the entire suite of kernel methods for the study of FCMs to facilitate visualization, classification, and general pattern recognition. 

In conclusion, this study has provided insights into the evaluation of LLM-generated FCMs and also underscores the need for continued research. Our framework provides a structured approach for these evaluations. This paper marks just the beginning of a journey towards improving the overall evaluation framework for FCMs and enhancing the role of LLMs in collective intelligence research, particularly in SES contexts with small quantities of low quality textual data.

\section*{Acknowledgments}

We thank Krithi Sachithanand, Anna Jeffries, and Stephen Lee for input during the exploratory phase of this project. We express deep gratitude to MSU professors Dirk Colbry, Paul Speaker, and Steven A. Gray as well as Dr. Holden Harris and Willem Kajbor of NOAA for enabling this research collaboration. SD was supported by NSF grant RC112900, Collaborative Research: Integrating Perspective-taking and Systems Thinking for Complex Problem-Solving. NB and SLG were supported by the National Oceanic and Atmospheric Administration's RESTORE Science Program under award NA22NOS4510203 to the National Center for Ecological Analysis and Synthesis at the University of California, Santa Barbara as part of the Offshore Wind \& Fisheries Gulf Ecosystem Initiative. NB, SLG, and SEF were supported by the Army Contracting Command, DARPA, and ARO under contract no. W911NF-21-C-0007. The views, opinions and/or findings expressed are those of the authors and should not be interpreted as representing the official views or policies of the Department of Defense or the U.S. Government.

\bibliography{acl_latex}

\appendix



\section{Limitations}
While our approach is robust, it is important to acknowledge potential limitations that could impact the generalizability and effectiveness of our findings.

Firstly, the passages we selected for our study are specific to a particular context and may not be representative of different domains, which could limit the generalizability of our findings across other contexts. Future research should explore cross-domain evaluations to validate the effectiveness of our approach in various settings.

Secondly, our methodology relies on initial human annotations, then selecting between human and LLM annotations as the gold standard, which introduces the possibility of bias due to the diversity of cultural and disciplinary backgrounds of the annotators. The initial annotations may not encompass all possible interpretations or nuances present in the text. Future work will focus on expanding the dataset to include a wider variety of texts and annotations, which will help in creating a more comprehensive and representative gold standard. 

Furthermore, while we aimed to fine-tune LLMs for improved performance, we did not tune all hyperparameters. Specifically, we only optimized the rank parameter $r$ for LoRA. The primary reason for not extensively tuning all hyperparameters, such as the learning rate, was that the focus of this paper was on measure alignment instead of optimizing hyperparameter settings. Future studies should aim to explore a broader range of hyperparameter tuning to fully explore the capabilities of the LLMs.

Additionally, our experiment was conducted with a limited number of annotation samples for LLM training and Elo ranking. Although the sample selection aimed to cover a broad spectrum of text complexities, the small sample size may not fully capture the variability in real-world data. Furthermore, the samples were selected to provide difficult examples, which may not represent typical data. Moreover, the limited sample size may limit the capabilities of LLMs due to a lack of surplus of data available for fine tuning. Elo rankings may have marginally deviated due to a limited sample size. Expanding the number of samples in future experiments will enhance the reliability and applicability of our results.

Moreover, our current approach does not leverage human-in-the-loop (HITL) strategies to iteratively improve LLM inferences based on human feedback. Integrating HITL mechanisms with the Elo rating system could significantly enhance the quality and accuracy of LLM-generated annotations. By continuously integrating human judgment, this iterative process would allow for ongoing refinement and improvement of LLM outputs. Future work should explore implementing HITL strategies to capture real-time human feedback and use it to fine-tune and validate LLM performance.

To summarize, our approach demonstrates promise. However, addressing these limitations in future work will be crucial for further validating and enhancing the robustness, reliability, and applicability of our findings. 

\section{Ethical Considerations}

This study involves the extraction and validation of fuzzy cognitive maps (FCMs) from text using large language models (LLMs). Several ethical considerations are relevant to this work, particularly regarding data use, annotation processes, biases, and the environmental impact of our research.

\textbf{Data Use and Privacy:} The data used in this study were created and annotated by the authors. This ensures that we have complete control over the data's provenance and the conditions under which it was generated. Since the data were produced specifically for this research, issues related to intellectual property and participant privacy are minimized.

\textbf{Annotation Process:} All annotations were carried out by the authors, ensuring a consistent understanding of the task and eliminating the need for external annotators. This method addresses concerns about fair compensation and working conditions for annotators, as the work was part of the authors' research activities.

\textbf{Bias and Fairness:} Inherent biases in language models can affect the outcomes. Researchers should explore methods to identify and mitigate such biases to enhance the fairness and reliability of FCM extractions.

\textbf{Environmental Impact:} The environmental impact of training and fine-tuning LLMs is a significant concern in NLP research. In our study, each training session lasted approximately 40 minutes, which is relatively short. This brevity was due to our primary focus on developing and validating measures for extracting and evaluating FCMs from text, rather than optimizing LLM performance. Consequently, we did not extensively tune the LLM hyperparameters, such as the learning rate, as our focus was on measure alignment rather than finding the ideal hyperparameter settings. This approach not only aligns with our research goals but also minimizes the environmental footprint of our computational experiments.

\textbf{Potential Misuse:} NLP technologies can be misused in various ways, such as generating misleading information or reinforcing harmful stereotypes. Researchers and practitioners should be aware of these risks and take steps to mitigate them when deploying such technologies.

Researchers should incorporate comprehensive strategies to address these ethical challenges, ensuring that the development and application of NLP technologies are aligned with broader societal values and ethical standards.

\section{Fine-Tuning Parameters and Hyperparameters}
\label{parameters}
For fine-tuning the models with instruction tuning, we focused on adjusting the rank $r$ in LoRA, while maintaining other training parameters at constant values. The cost function for training and validation was cross-entropy loss. The Huggingface library \cite{huggingface} was utilized to run the training jobs with 4-bit quantization.

The common hyperparameters and their corresponding values used for fine-tuning the three models are listed below. Note that while the maximum number of training epochs was set to 15, early stopping was employed, so not all experiments reached the full 15 epochs. The early stopping mechanism halted training when the validation loss did not improve for 3 consecutive epochs.

\begin{itemize}[noitemsep]
\item \texttt{Maximum number of training epochs: 15 (subject to early stopping)}
\item \texttt{Batch size: 4}
\item \texttt{Optimizer: Paged AdamW 32-bit}
\item \texttt{Learning rate: 2e-4}
\item \texttt{Learning rate scheduler: Cosine decay}
\item \texttt{Gradient accumulation steps: 1}
\item \texttt{Gradient clipping: 0.3}
\item \texttt{Gradient checkpointing : True (to save memory)}
\item \texttt{Weight decay: 0.001}
\item \texttt{Warmup ratio: 0.1}
\item \texttt{Use of 4-bit precision: Enabled (to reduce memory and computational cost)}
\item \texttt{Data type for 4-bit computations: bfloat16}
\item \texttt{Quantization type for 4-bit precision: nf4}
\item \texttt{Nested quantization: Disabled}
\item \texttt{LoRA dropout rate: 0.1}
\end{itemize}

The following hyperparameters were optimized during the fine-tuning process:

\begin{itemize}
\item \texttt{LoRA rank (r): 2, 4, 8, 16, 32, 64, 128, 256}
\item \texttt{LoRA $\alpha: 2 * r$}
\end{itemize}

We used one Nvidia V100 GPU to execute the training jobs. On average, each experiment took approximately 40 minutes to complete.

\subsection{Optimal Rank (r) Values for LoRA Fine-Tuning}
\label{ranks}

To determine the optimal rank $r$ for each model, we experimented with various $r$ values and monitored the validation loss.  

The best $r$ values, based on the minimum validation set loss for Llama-2-7B-chat-hf, Llama-3-8B-Instruct, and Mistral-7B-Instruct-v0.2 after testing different $r$ values, are as follows:

\begin{itemize}[noitemsep]
\item \texttt{Llama-2-7B-chat-hf: 128}
\item \texttt{Llama-3-8B-Instruct: 64}
\item \texttt{Mistral-7B-Instruct-v0.2: 128}
\end{itemize}

\begin{figure*}[h]
\centering
\includegraphics[width=0.9\linewidth]{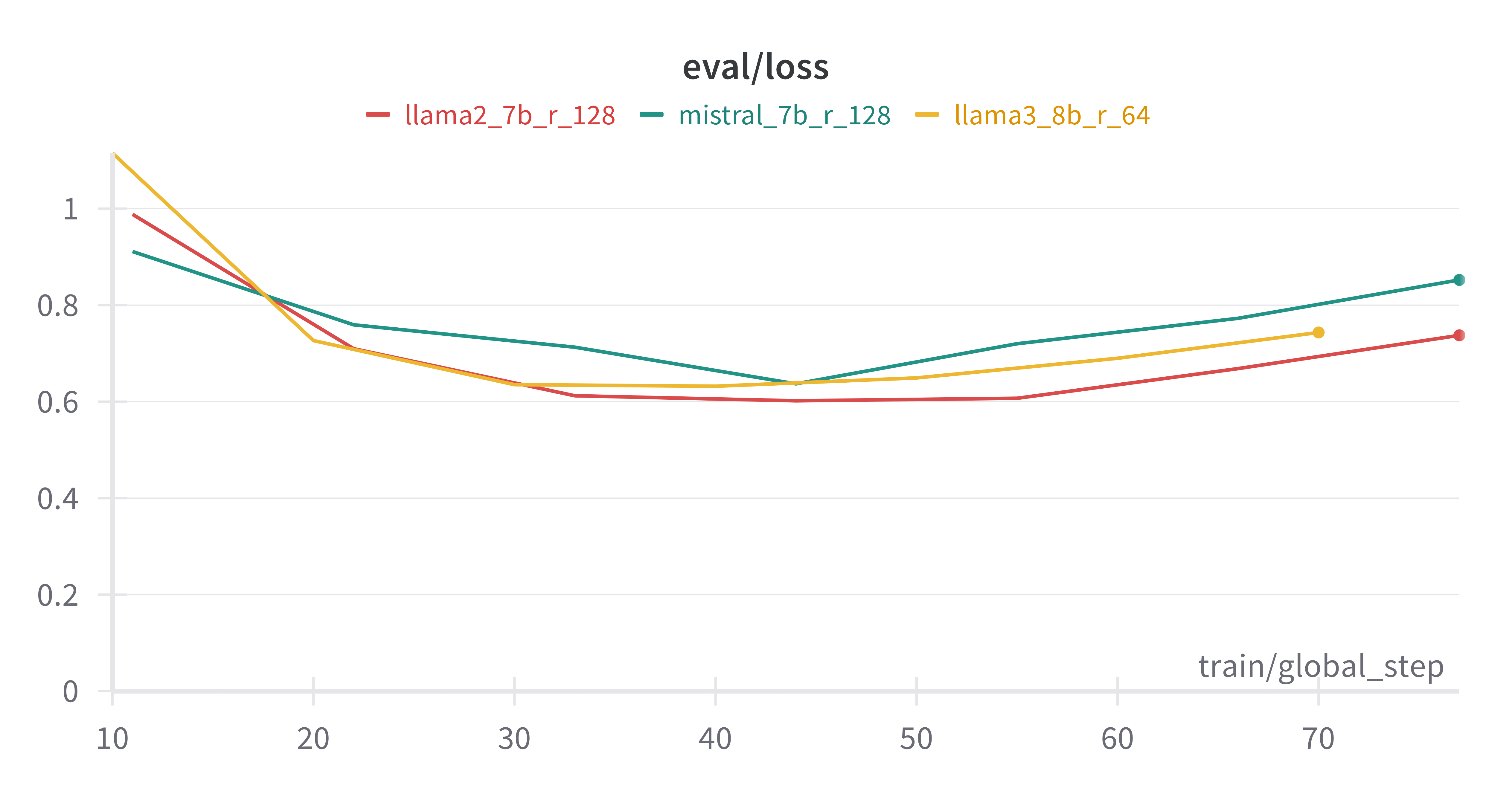}
\caption{Validation losses for Mistral-7B-Instruct-v0.2, Llama-2-7B-chat-hf, and Llama-3-8B-Instruct during training with their respective optimal ranks $r$ in LoRA. Plot generated using WandB \cite{wandb}.}
\label{fig:ranks}
\end{figure*}

As shown in Figure \ref{fig:ranks}, all three models achieved similar validation losses with their respective optimal $r$ values.

\section{Other Formulas and Hyperparameters}
\subsection{Elo Rating System}
The Elo rating system \cite{elo, eloBook}, is a method for quantifying the relative skill levels of players in two-player competitive games such as chess. It assigns a numerical rating to each player, representing their skill level.

The Elo rating system updates players' ratings after each game based on the outcome. The formula to update the rating involves several steps. First, the expected score for player $A$ against player $B$ is calculated using the formula:
\begin{equation*}
\label{eq:EA}
E_A=\frac{1}{1+10^{(R_B-R_A)/400}},
\end{equation*}
where $R_A$ and $R_B$ are the current ratings of players $A$ and $B$, respectively. Similarly, the expected score for player B is:
\begin{equation*}
\label{eq:EB}
E_B=\frac{1}{1+10^{(R_A-R_B)/400}}.
\end{equation*}
Note that $E_A+E_B=1$. The actual score $S_A$ is 1 if player $A$ wins, 0 if player $A$ loses, and 0.5 in the case of a draw. Similarly, $S_B$ is 1 if player $B$ wins, 0 if player $B$ loses, and 0.5 for a draw. The new ratings for players $A$ and $B$ are updated using the formulas:
\begin{equation*}
\label{eq:RA}
\begin{array}{l}
R'_A = R_A + K(S_A - E_A)  \\ 
R'_B = R_B + K(S_B - E_B)
\end{array}
\end{equation*}
Here, $K$ is a constant known as the $K-$factor, which determines the sensitivity of the rating system. A higher $K-$factor means ratings change more significantly after each game, making the system more responsive to recent results. Conversely, a lower $K$–factor results in smaller changes, making the ratings more stable and less sensitive to new results. The $K-$factor can vary; in chess, it is often set to 32 for new players and 16 for established players, but these values can be adjusted depending on the specific application and the desired stability of the ratings.

It can be shown that while the values of $K$ and the initial Elo ratings $R_A$ and $R_B$ affect the numerical scores obtained after calculation, the \textit{relative ranking} of players remains unchanged. Additionally, the order of "games" does not affect the final ranking, ensuring the consistency of the system regardless of the sequence in which games are played.

\subsection{Hyperparameters for Elo Rating Calculation}

In the process of calculating Elo ratings for annotation evaluations, we utilized the following hyperparameters:

\begin{itemize}[noitemsep]
\item \texttt{K–factor: 32}
\item \texttt{Initial Elo rating: 1000}
\end{itemize}


\subsection{Hyperparameters for Similarity Measures}\label{app:similarity-hyperparams}
The English-trained checkpoint bleurt-base-128 and ROUGE-1 were used in this study. We considered a range of thresholds, $T$, for each measure through exploratory data analysis and adaptive grid search. The $T$ chosen for each measure coordinates to the highest achieved Spearman correlation to the human-generated rankings. The $T$  selected for each measure is:

\begin{itemize}[noitemsep]
\item \texttt{BLEURT-E: -0.1532}
\item \texttt{BLEU-E: 0.352}
\item \texttt{METEOR-E: 0.01}
\item \texttt{ROUGE-E: 0.45}.
\end{itemize}

\section{Prompts Format for Instruction Tuning}
\label{prompt}

This section details the prompts format used for instruction tuning. This format ensures that the model clearly understands the task and generates the appropriate response based on the given instruction.

\subsection{Llama-2-7B and Mistral-7B}
Both Llama-2-7B-chat-hf and Mistral-7B-Instruct-v0.2 use the same prompt format for instruction tuning. We provide the instructions for the model within the \verb|[INST]| and \verb|[/INST]| tags, and the model generates everything following the \verb|[/INST]| tag. The entire prompt is enclosed within \verb|<s>| and \verb|</s>| tags.

The prompt format used is mentioned below:

\small
\begin{lstlisting}
<s>[INST] Given the input sentence, identify all the triplets of entities and the corresponding causal relationships between them. The entities should be phrases from the input sentence, and the relationships should either be 'Positive' or 'Negative'. Each new extracted triplet should start with the <triplet> token, followed by the subject phrase, the object phrase, and the relationship, separated by <subj> and <obj> tokens.

Input Sentence: <Sentence> [/INST]

Causal Relation Triplets : <triplet> Subject_1 <subj> Target_1 <obj> Relationship_1 
<triplet> Subject_2 <subj> Target_2 <obj> Relationship_2 </s>
\end{lstlisting}
\normalsize

A complete example, including the prompt, a sample sentence, and its causal relation triplets used for instruction tuning, is provided below:

\small
\begin{lstlisting}
<s>[INST] Given the input sentence, identify all the triplets of entities and the corresponding causal relationships between them. The entities should be phrases from the input sentence, and the relationships should either be 'Positive' or 'Negative'. Each new extracted triplet should start with the <triplet> token, followed by the subject phrase, the object phrase, and the relationship, separated by <subj> and <obj> tokens.

Input Sentence: Islamist violence in Mali has also hit cattle herding areas, forcing farmers to abandon their trade. Climate change too has led to competition for grazing lands and water, leading to intercommunal conflicts. The result, increased costs for breeders. [/INST]

Causal Relation Triplets : <triplet> islamist violence <subj> cattle herding <obj> negative 
<triplet> climate change <subj> competition for grazing lands and water <obj> positive 
<triplet> competition for grazing lands and water <subj> intercommunal conflicts <obj> positive 
<triplet> intercommunal conflicts <subj> increased costs for breeders <obj> positive </s>
\end{lstlisting}
\normalsize

\subsection{Llama-3-8B}
Llama-3-8B-Instruct follows a different prompt format compared to Llama-2 or Mistral. An example of the prompt format for Llama-3-8B-Instruct is provided below:

\small
\begin{lstlisting}
<|begin_of_text|><|start_header_id|> system<|end_header_id|>

Given the input sentence, identify all the triplets of entities and the corresponding causal relationships between them. The entities should be phrases from the input sentence, and the relationships should either be 'Positive' or 'Negative'. Each new extracted triplet should start with the <triplet> token, followed by the subject phrase, the object phrase, and the relationship, separated by <subj> and <obj> tokens. <|eot_id|><|start_header_id|> user<|end_header_id|>

Input Sentence : A direct negative effect that for example a wind farm can have on the trawl fishery (reduced fishing activity), <|eot_id|><|start_header_id|> assistant<|end_header_id|>

Casual Relation Triplets : <triplet> wind farm <subj> trawl fishery <obj> negative <|eot_id|>

\end{lstlisting}
\normalsize

\section{Prompts Format For Zero- and Three-Shot Learning}
\label{zero_three_shot_prompt}

\subsection{Zero-Shot Learning}
\label{zeroshotapp}

To achieve structured output in zero-shot and three-shot in-context learning, we optimized the prompts. Examples of the prompts used in zero-shot in-context learning for all the models are as follows:

\subsubsection{Llama-2-7B-chat-hf}
\small
\begin{lstlisting}
    <s>[INST] <<SYS>> Given the input sentence, identify all the triplets (subject, object and causal relation) . The subject and object should be phrases from the input sentence. 
    The causal relation between subject and object should strictly be either "Positive" or "Negative" and nothing else. 
    Each new extracted triplet i.e. subject, object and relation should start with a newline should be within <triple> and </triplet>. The subject should be within <subj> and </subj> tokens. The object should be within <obj> and </obj> tokens. The causal relation should be within <relation> and </relation> tokens. The format of output of each triplet should be strictly like below:
    
    <triplet>
        <subj> </subj>
        <obj> </obj>
        <relation> </relation>
    </triplet>
    <</SYS>>
    Input Sentence : pastoralists in the arid and semi-arid regions of Mali continue to face increasing risk due to low levels of rainfall [/INST]
    Causal Relation Triplet : 
        
    <triplet>
        <subj> pastoralists</subj>
        <obj> low levels of rainfall</obj>
        <relation> Negative</relation>
    </triplet>
    
    <triplet>
        <subj> Mali</subj>
        <obj> increasing risk</obj>
        <relation> Positive</relation>
    </triplet>
    
    Note: The subject and object phrases are enclosed within <subj> and <obj> tokens, respectively, and the causal relation is enclosed within <relation> and </relation> tokens.
\end{lstlisting}
\normalsize

\subsubsection{Llama-3-8B-Instruct}
\small
\begin{lstlisting}
    <|begin_of_text|> <|start_header_id|> system <|end_header_id|>Given the input sentence, identify all the triplets (subject, object and causal relation). The subject and object should be phrases from the input sentence. 
    The causal relation between subject and object should strictly be either "Positive" or "Negative" and nothing else. 
    Each new extracted triplet i.e. subject, object and relation should start with a newline should be within <triple> and </triplet>. The subject should be within <subj> and </subj> tokens. The object should be within <obj> and </obj> tokens. The causal relation should be within <relation> and </relation> tokens.
    The format of output of each triplet should be strictly like below:
    <triplet>
        <subj> </subj>
        <obj> </obj>
        <relation> </relation>
    </triplet> <|eot_id|><|start_header_id|> user <|end_header_id|>
    Input Sentence : Women identified forced sex and men highlighted lack of awareness about contraception and fear of side effects as important causes of kunika. <|eot_id|><|start_header_id|> assistant <|end_header_id|>
    <triplet>
        <subj>Women</subj>
        <obj>forced sex</obj>
        <relation>Positive</relation>
    </triplet>
    
    <triplet>
        <subj>men</subj>
        <obj>lack of awareness about contraception and fear of side effects</obj>
        <relation>Positive</relation>
    </triplet>
    
    <triplet>
        <subj>men</subj>
        <obj>kunika</obj>
        <relation>Negative</relation>
    </triplet>
\end{lstlisting}
\normalsize

\subsubsection{ Mistral-7B-Instruct-v0.2}
\small
\begin{lstlisting}
    <s>[INST] Given the input sentence, identify all the triplets (subject, object and causal relation) . The subject and object should be phrases from the input sentence. 
    The causal relation between subject and object should strictly be either "Positive" or "Negative" and nothing else. 
    Each new extracted triplet i.e. subject, object and relation should start with a newline should be within <triple> and </triplet>. The subject should be within <subj> and </subj> tokens. The object should be within <obj> and </obj> tokens. The causal relation should be within <relation> and </relation> tokens. The format of output of each triplet should be strictly like below:
    <triplet>
        <subj> </subj>
        <obj> </obj>
        <relation> </relation>
    </triplet>

    Input Sentence : pastoralists in the arid and semi-arid regions of Mali continue to face increasing risk due to low levels of rainfall [/INST]
    Causal Relation Triplet : 
     <triplet>
        <subj> pastoralists in the arid and semi-arid regions of Mali </subj>
        <obj> face increasing risk </obj>
        <relation> Positive </relation>
    </triplet>
    <triplet>
        <subj> Low levels of rainfall </subj>
        <obj> cause pastoralists in the arid and semi-arid regions of Mali to face increasing risk </obj>
        <relation> Negative </relation>
    </triplet>
\end{lstlisting}
\normalsize

\subsection{Three-Shot Learning}
\label{threeshotapp}

To achieve structured output in three-shot in-context learning, we optimized the prompts. Below are examples of the prompts used in three-shot in-context learning for all the models: 

\subsubsection{Llama-2-7B-chat-hf}
\small
\begin{lstlisting}
    <s>[INST] <<SYS>> Given the input sentence, identify all the triplets of entities and the corresponding causal relationships between them. The entities should be phrases from the input sentence, and the relationships should either be 'Positive' or 'Negative'. Each new extracted triplet should start with the <triplet> token, followed by the subject phrase, the object phrase, and the relationship, separated by <subj> and <obj> tokens.
    Don't add extra sentences.
    <</SYS>>
    Input Sentence : the current price of local rice (sold loose) at the local market is 1850 ngn/1kg. the price is expected to rise to 2100 ngn/1kg in 6 weeks, due to the high cost of oil. [/INST]
    Causal Relation Triplets : <triplet> high cost of oil <subj> price of local rice <obj> positive
    </s>
    [INST]
    Input Sentence : Participants also believed that illiteracy and low levels of education among some of the women were barriers to seeking skilled pregnancy health care. [/INST]
    Causal Relation Triplets : <triplet> illiteracy among women <subj> access to skilled pregnancy health care <obj> negative 
    <triplet> low education among women are understaffed <subj> access to skilled pregnancy health care <obj> negative
    </s>
    [INST]
    Input Sentence : Other health sources of protein are lean meats, low-fat milk, nuts, and beans such as kidney beans. [/INST]
    Causal Relation Triplets : <triplet> nuts <subj> health sources of protein <obj> positive 
    <triplet> meats <subj> health sources of protein <obj> positive </s>
    <triplet> milk <subj> health sources of protein <obj> positive </s>
    <triplet> beans <subj> health sources of protein <obj> positive </s>
    </s>
    [INST]
    Input Sentence : pastoralists in the arid and semi-arid regions of Mali continue to face increasing risk due to low levels of rainfall [/INST]
    Causal Relation Triplets : 
     <triplet> low rainfall <subj> risk faced by pastoralists <obj> negative
\end{lstlisting}
\normalsize

\subsubsection{Llama-3-8B-Instruct}
\small
\begin{lstlisting}
    <|begin_of_text|> <|start_header_id|> system <|end_header_id|> Given the input sentence, identify all the triplets of entities and the corresponding causal relationships between them. The entities should be phrases from the input sentence, and the relationships should either be 'Positive' or 'Negative'. Each new extracted triplet should start with the <triplet> token, followed by the subject phrase, the object phrase, and the relationship, separated by <subj> and <obj> tokens.
    Don't add extra sentences. <|eot_id|><|start_header_id|> user <|end_header_id|>
    Input Sentence : the current price of local rice (sold loose) at the local market is 1850 ngn/1kg. the price is expected to rise to 2100 ngn/1kg in 6 weeks, due to the high cost of oil. <|eot_id|><|start_header_id|> assistant <|end_header_id|>
    Causal Relation Triplets : <triplet> high cost of oil <subj> price of local rice <obj> positive <|eot_id|><|start_header_id|> user <|end_header_id|>

    Input Sentence : Participants also believed that illiteracy and low levels of education among some of the women were barriers to seeking skilled pregnancy health care. <|eot_id|><|start_header_id|> assistant <|end_header_id|>
    Causal Relation Triplets : <triplet> illiteracy among women <subj> access to skilled pregnancy health care <obj> negative 
    <triplet> low education among women are understaffed <subj> access to skilled pregnancy health care <obj> negative <|eot_id|><|start_header_id|> user <|end_header_id|>

    Input Sentence : Other health sources of protein are lean meats, low-fat milk, nuts, and beans such as kidney beans.<|eot_id|><|start_header_id|> assistant <|end_header_id|>
    Causal Relation Triplets : <triplet> nuts <subj> health sources of protein <obj> positive 
    <triplet> meats <subj> health sources of protein <obj> positive 
    <triplet> milk <subj> health sources of protein <obj> positive 
    <triplet> beans <subj> health sources of protein <obj> positive <|eot_id|><|start_header_id|> user <|end_header_id|>

    Input Sentence : pastoralists in the arid and semi-arid regions of Mali continue to face increasing risk due to low levels of rainfall <|eot_id|><|start_header_id|> assistant <|end_header_id|>
   Causal Relation Triplets : <triplet> forced sex <subj> kunika <obj> positive 
 <triplet> lack of awareness about contraception <subj> kunika <obj> positive 
 <triplet> fear of side effects <subj> kunika <obj> positive
\end{lstlisting}
\normalsize

\subsubsection{Mistral-7B-Instruct-v0.2}
\small
\begin{lstlisting}
    <s>[INST] Given the input sentence, identify all the triplets of entities and the corresponding causal relationships between them. The entities should be phrases from the input sentence, and the relationships should either be 'Positive' or 'Negative'. Each new extracted triplet should start with the <triplet> token, followed by the subject phrase, the object phrase, and the relationship, separated by <subj> and <obj> tokens.
    Don't add extra sentences.
    Input Sentence : the current price of local rice (sold loose) at the local market is 1850 ngn/1kg. the price is expected to rise to 2100 ngn/1kg in 6 weeks, due to the high cost of oil. [/INST]
    Causal Relation Triplets : <triplet> high cost of oil <subj> price of local rice <obj> positive
    </s>
    [INST]
    Input Sentence : Participants also believed that illiteracy and low levels of education among some of the women were barriers to seeking skilled pregnancy health care. [/INST]
    Causal Relation Triplets : <triplet> illiteracy among women <subj> access to skilled pregnancy health care <obj> negative 
    <triplet> low education among women are understaffed <subj> access to skilled pregnancy health care <obj> negative
    </s>
    [INST]
    Input Sentence : Other health sources of protein are lean meats, low-fat milk, nuts, and beans such as kidney beans. [/INST]
    Causal Relation Triplets : <triplet> nuts <subj> health sources of protein <obj> positive 
    <triplet> meats <subj> health sources of protein <obj> positive </s>
    <triplet> milk <subj> health sources of protein <obj> positive </s>
    <triplet> beans <subj> health sources of protein <obj> positive </s>
    </s>
    [INST]
    Input Sentence : pastoralists in the arid and semi-arid regions of Mali continue to face increasing risk due to low levels of rainfall [/INST]
    Causal Relation Triplets : 
     <triplet> low levels of rainfall <subj> risk faced by pastoralists in arid and semi-arid regions of Mali <obj> positive. 
\end{lstlisting}
\normalsize

\section{User Interfaces}
\label{app:UI}

In this appendix, we provide screenshots of the two custom Dash-based user interfaces (UIs) \cite{dash} developed for this study. These UIs were integral to the annotation and evaluation processes, facilitating consistent data collection and pairwise comparisons.

\subsection{Annotation Interface}

\begin{figure*}[h]
    \centering
   \includegraphics[width=1\textwidth]{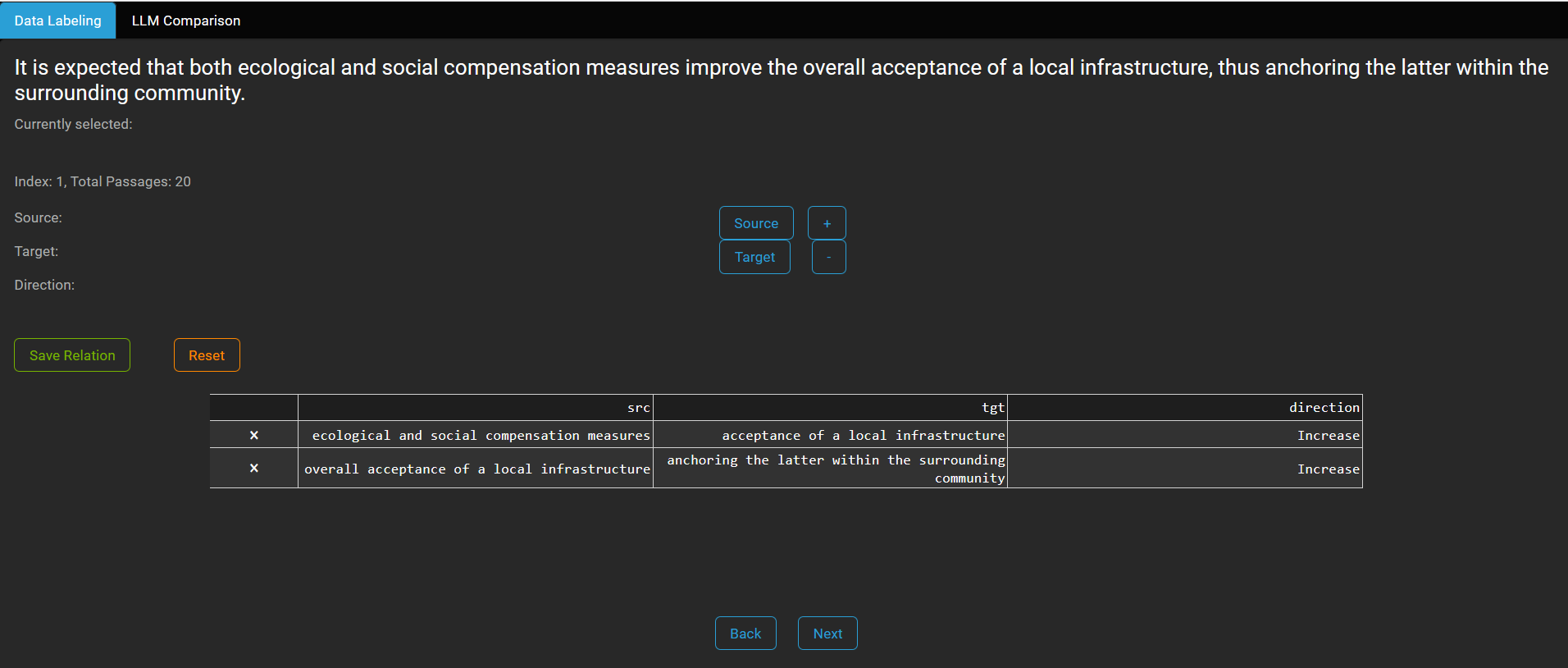}
    \caption{Screenshot of the annotation interface used by participants to annotate the text samples.}
    \label{fig:annotation}
\end{figure*}

Figure \ref{fig:annotation} shows the UI used by the seven participants to annotate the 20 selected samples. This interface was designed to be user-friendly and efficient, allowing participants to focus on the quality of their annotations. The annotations were saved in JSON format for consistency and ease of processing.

\subsection{Elo Comparison Interface}

\begin{figure*}[h]
    \centering
    \includegraphics[width=1\textwidth]{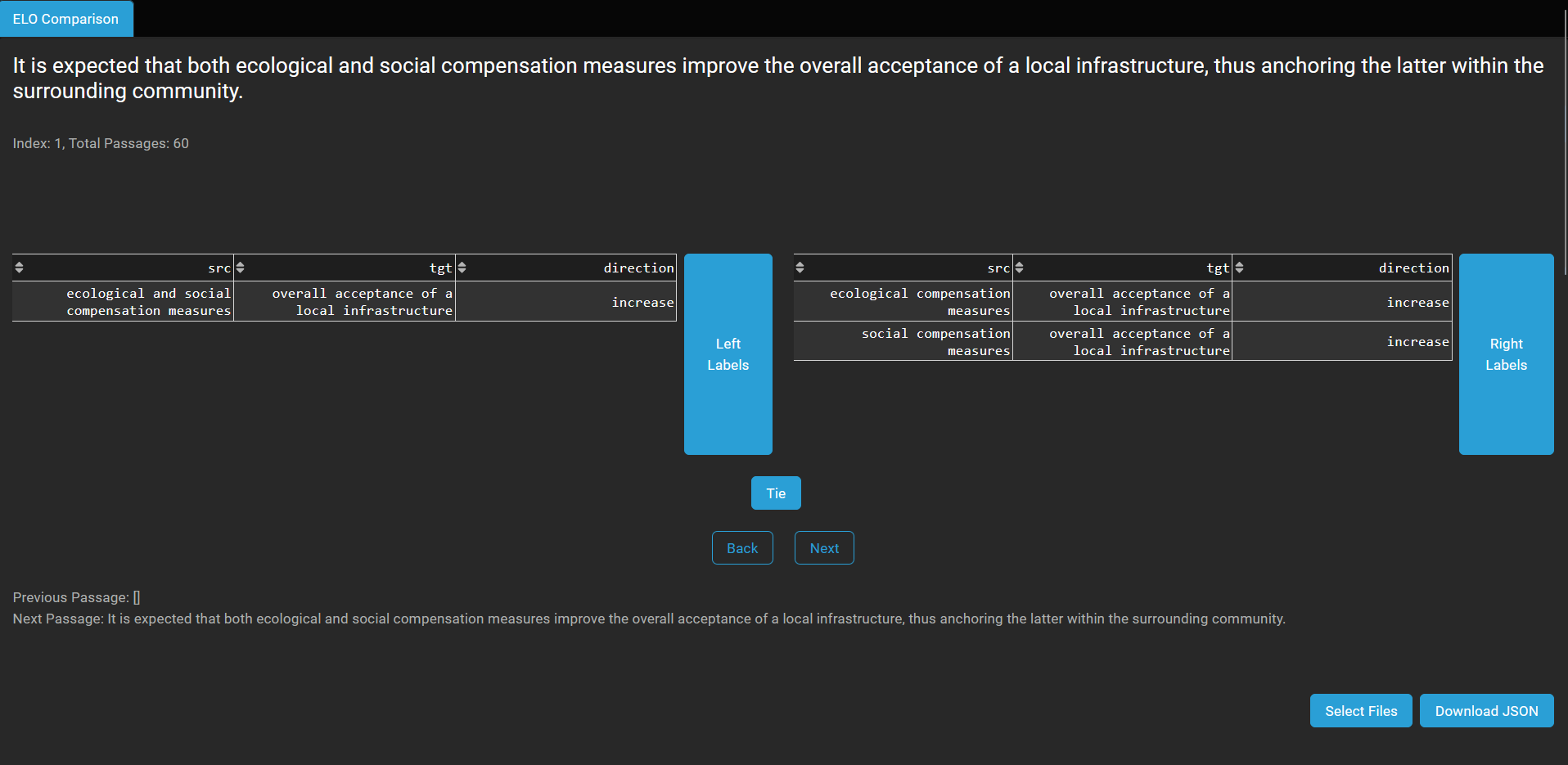}
    \caption{Screenshot of the Elo comparison interface used for pairwise comparisons of annotations. This interface helped in generating the Elo scores by allowing participants to select the better annotation in each presented pair.}
    \label{fig:elo}
\end{figure*}

Figure \ref{fig:elo} displays the UI used for the Elo rating comparisons. In this interface, pairs of annotations were presented to participants, who were asked to choose the better annotation for each pair. This interface randomized the sequence of comparisons to eliminate potential biases and ensured that participants could not see their own annotations to prevent bias.

\section{Split Creation}
\label{elo splits}
The splits were created by taking the labelers annotations (7), extracting the names into a set, and generating a list of random numbers the length of which is the amount of combinations times the amount of texts chosen. Then, with nested loops, the outer of which being the texts, and the inner being the possible labeler combinations, these numbers were assigned in order (front of list to back of list). The dictionary was then sorted and made back into a dictionary. Then, this list was divided into splits by iterating through it, assigning one datapoint at a time to each labeler skipping that point if the labeler that is up is in the combinations. This loop continues until all points are assigned.

LLM additions followed a similar process. We looped through each labeler and created a combination list with them and the LLMs, but not the LLMs with each other. After, we generated values a random list and added the current length of the list to each point. Then, looping through the rest of the labelers, we follow the same process of looping through the dictionary and assigning one point at a time to a labeler. After all labelers had been paired with the LLMs, we created the combination list of the LLMs with each other, and followed the same process of looping through the labelers.

Finally, we generated a list of random numbers for each labeler as long as an individual labeler’s split. Then, we reassigned the key values to the new list of random numbers to obscure the ordering in which the labels were added so as to randomize the order of presentation of pairings between labeler+labeler, labeler+LLM and LLM.

Inter-rater splits were created using only LLM outputs, where each labeler compared the three combinations for each of 20 samples.

\subsection{Inter- and Intra-Rater Reliability}\label{app:inter-rater}

\begin{figure}[h]
    \centering
    \includegraphics[width=\linewidth]{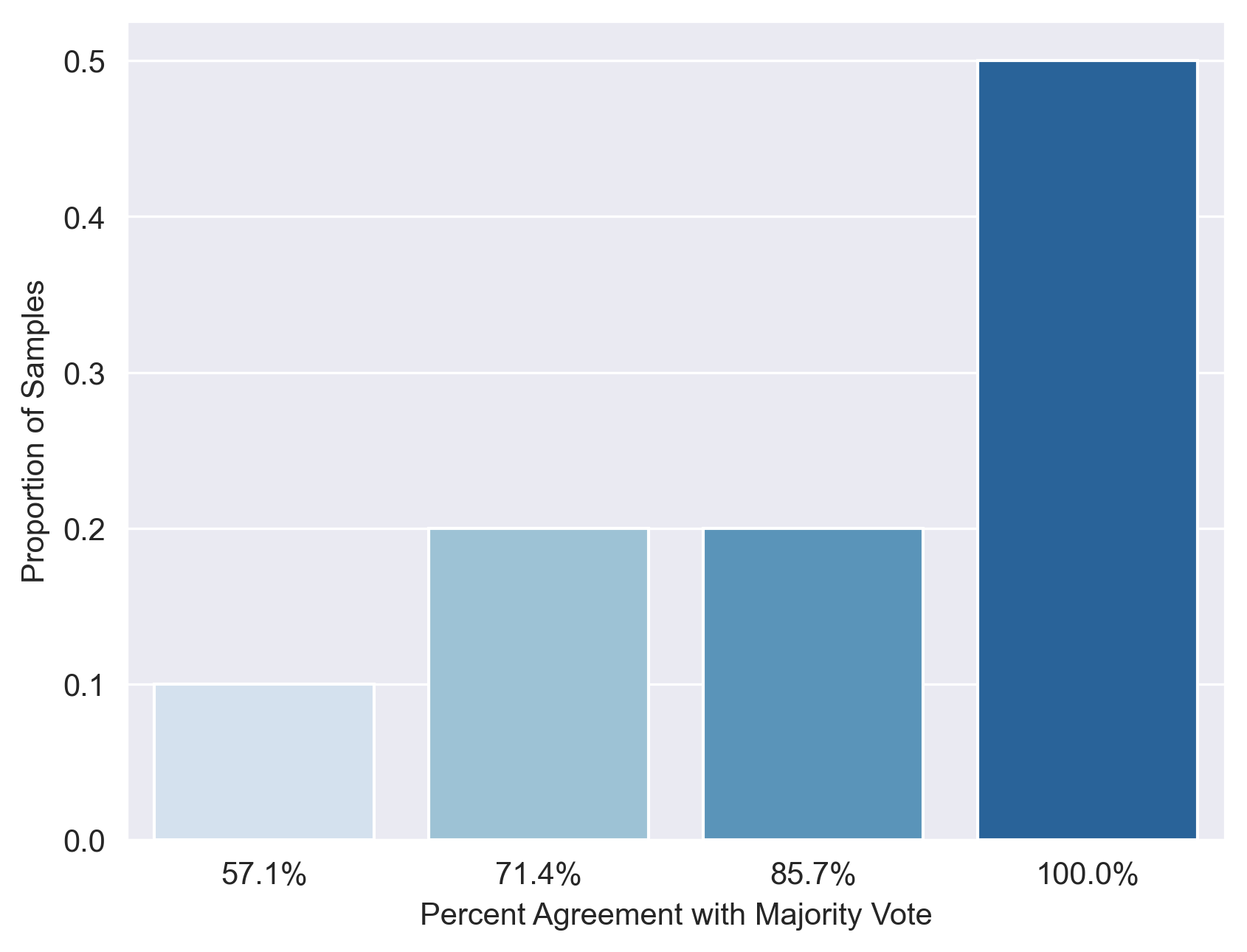}
    \caption{Inter-rater reliability for paired annotations}
    \label{fig:inter-rater}
\end{figure}

The bar plot in Figure \ref{fig:inter-rater} illustrates the inter- and intra-rater reliability among different percentages of raters. Specifically, 57.1\% of raters agreed on 10\% of the samples, 71.4\% of raters agreed on 20\% of the samples, another 85.7\% of raters also agreed on 20\% of the samples, and finally, 100\% of raters agreed on 50\% of the samples. Notably, this distribution shows that 90\% of the data (0.2 + 0.2 + 0.5) received agreement from 71.4\% of raters or more, highlighting a substantial consensus among the majority of raters in this evaluation. Raters, furthermore, demonstrated 90.5\% self-consistency.

\section{Rater Guidelines}
\label{app:rater-guidelines}
To support annotation scoring consistency and scientific reproducibility, raters were instructed to use the following guidelines when choosing a winner during pairwise comparisons:
\begin{enumerate}
    \item Prefer more `better' tuples rather than more `worse' ones (this constitues our 'human-thresholded' F1 per discussion today).
    \item Prefer node names that do not introduce new concepts that are not present in the text.
    \item Prefer source/target in correct position.
    \item If A affects B and if B affects C, we can also infer that A affects C. However, we should not include "A affects C'' in the annotations, unless it is explicitly mentioned.
    \item Prefer node names as close to the text as possible.
    \item Prefer verbose node names (include adjectives) as these contain information that can be referenced, utilized, or abstracted away as necessary by downstream tasks/processes.
    \item Prefer splitting node names at `and' conjunctions when concepts are `distinct', e.g., in ``It is expected that both ecological and social compensation measures improve the overall acceptance of a local infrastructure, thus anchoring the latter within the surrounding community'' it is better to have one source nodes corresponding to \textit{ecological compensation measures} as well as one corresponding to \textit{social compensation measures}.
    \item Prefer correct direction of causal relation.
\end{enumerate}
Some of these guidelines can be in conflict with each other. In these cases, prefer a heuristic higher on the above list; but in all cases, use your best judgment.

\section{Contrasts with F1}
Table \ref{table:spearman_diff_results} provides paired differences in correlations between edge-based measures and vanilla F1 rankings. 
\begin{table}[h]
\centering
\footnotesize
\begin{tabular}{l l l l}
\toprule
\textbf{Measure} & \textbf{Mean} & \textbf{90\% CI} & \textbf{95\% CI} \\ \toprule
BLEU-E &\textbf{0.399} & (\textbf{0.263}, \textbf{ 0.535}) & (\textbf{0.234}, \textbf{0.564}) \\ 
METEOR-E & \textbf{0.317} & (\textbf{0.142}, \textbf{0.492}) & (\textbf{0.105}, \textbf{0.528}) \\ 
ROUGE-E & \textbf{0.371} & (\textbf{0.210}, \textbf{0.532}) & (\textbf{0.176}, \textbf{0.566}) \\ 
BLEURT-E & \textbf{0.322} & (\textbf{0.173}, \textbf{0.471}) & (\textbf{0.141}, \textbf{0.503}) \\  \bottomrule
\end{tabular}
\caption{Mean paired differences between similarity measures' and the baseline F1 measure's correlations.}
\label{table:spearman_diff_results}
\end{table}

\section{Datasheet}
\subsection{Motivation for Dataset Creation} 
\begin{itemize}
    \item \textit{Why was the dataset created? (e.g., were there specific tasks in mind, or a specific gap that needed to be filled?) }The dataset is an amalgamation of SES literature that is relevant to the authors of this article and/or their collaborators. Each sub-dataset was created in the process of developing FCM models of the relevant system.
    \item \textit{What (other) tasks could the dataset be used for? Are there obvious tasks for which it should not be used?} Absolutely. We plan to further utilize this data in the construction of 'collective intelligence' models of these SES.
    \item \textit{Has the dataset been used for any tasks already? If so, where are the results so others can compare (e.g., links to published papers)?} No.
    \item \textit{Who funded the creation of the dataset? If there is an associated grant, provide the grant number. }The dataset creation was funded under the grants listed in the Acknowledgements section.
\end{itemize}

\subsection{Dataset Composition} 
\begin{itemize}
    \item  \textit{What are the instances? (that is, examples; e.g., documents, images, people, countries) Are there multiple types of instances? (e.g., movies, users, ratings; people, interactions between them; nodes, edges) Are relationships between instances made explicit in the data (e.g., social network links, user/movie ratings, etc.)? How many instances of each type are there?} Instances are text passage-tuple pairs that correspond to text data and associated concept pairs connected by causal relation edges. The dataset is apportioned as follows:
    \begin{itemize}
        \item Total unique in Training: 224
        \item Total unique in Validation : 38
        \item Total unique in Testing : 56
    \end{itemize}
    \item \textit{What data does each instance consist of? “Raw” data (e.g., unprocessed text or images)? Features/attributes? Is there a label/target associated with instances? If the instances are related to people, are subpopulations identified (e.g., by age, gender, etc.) and what is their distribution? }
    See above. Concepts can consist of anything that, roughly, 'qualitatively or quantitatively increases or decreases.’
    \item \textit{Is everything included or does the data rely on external resources? (e.g., websites, tweets, datasets) If external resources, a) are there guarantees that they will exist, and remain constant, over time; b) is there an official archival version. Are there licenses, fees or rights associated with any of the data? }The data does not rely on external resources.
    \item \textit{Are there recommended data splits or evaluation measures? (e.g., training, development, testing; accuracy/AUC)} When training on multiple models, it is important to use a consistent test set.
    \item \textit{What experiments were initially run on this dataset? Have a summary of those results and, if available, provide the link to a paper with more information here. Any other comments? } N/A
\end{itemize}

\subsection{Data Collection Process} 
\begin{itemize}
    \item \textit{How was the data collected? (e.g., hardware apparatus/sensor, manual human curation, software program, software interface/API; how were these constructs/measures/methods validated?)} Text passages were collected by extracting raw text from PDF documents and encoding them as strings. Annotations were assigned via the UI described in the main paper body and appendices.
    \item \textit{Who was involved in the data collection process? (e.g., students, crowdworkers) How were they compensated? (e.g., how much were crowdworkers paid?)} Only authors were involved in the data collection process. Funding consisted of salary and hourly pay.
    \item \textit{Over what time-frame was the data collected? Does the collection time-frame match the creation time-frame?}  3 years.
    \textit{How was the data associated with each instance acquired? Was the data directly observable (e.g., raw text, movie ratings), reported by subjects (e.g., survey responses), or indirectly inferred/derived from other data (e.g., part of speech tags; model-based guesses for age or language)? If the latter two, were they validated/verified and if so how?} The data was directly observed in text.
    \item \textit{Does the dataset contain all possible instances? Or is it, for instance, a sample (not necessarily random) from a larger set of instances? If the dataset is a sample, then what is the population? What was the sampling strategy (e.g., deterministic, probabilistic with specific sampling probabilities)? Is the sample representative of the larger set (e.g., geographic coverage)? If not, why not (e.g., to cover a more diverse range of instances)? How does this affect possible uses?} This dataset is a sample. The population consists of "all possible passage-tuple pairs." The dataset is representative in the same sense as data in other fine-tuning efforts (i.e., there is not a precisely meaningful sense of 'representativeness').
    \item \textit{Is there information missing from the dataset and why? (this does not include intentionally dropped instances; it might include, e.g., redacted text, withheld documents) Is this data missing because it was unavailable?} N/A
    \item \textit{Are there any known errors, sources of noise, or redundancies in the data?} Conversion of PDF to raw text can introduce errors, which we manually evaluated through samples. 
\end{itemize}

\subsection{Dataset Distribution}
\begin{itemize}
    \item \textit{How is the dataset distributed? (e.g., website, API, etc.; does the data have a DOI; is it archived redundantly?)} 
It has thus far been shared only within the research team.
    \item \textit{When will the dataset be released/first distributed? (Is there a canonical paper/reference for this dataset?)}The dataset will be released upon publication of the work.
    \item \textit{What license (if any) is it distributed under? Are there any copyrights on the data?}
The data will be openly available.

\item \textit{Are there any fees or access/export restrictions?} 
No.
\end{itemize}

\subsection{Dataset Maintenance}
\begin{itemize}    
\item \textit{Who is supporting/hosting/maintaining the dataset?}    
\item \textit{How does one contact the owner/curator/manager of the dataset (e.g. email address, or other contact info)?} 
The dataset may be requested from the main author.

\item \textit{Will the dataset be updated? How often and by whom? How will updates/revisions be documented and communicated (e.g., mailing list, GitHub)? Is there an erratum?} Potentially.
    \item \textit{If the dataset becomes obsolete how will this be communicated?} 
N/A
    \item \textit{Is there a repository to link to any/all papers/systems that use this dataset?} N/A
    \item \textit{If others want to extend/augment/build on this dataset, is there a mechanism for them to do so? If so, is there a process for tracking/assessing the quality of those contributions. What is the process for communicating/distributing these contributions to users?} 
Please send a request to the corresponding author.
\end{itemize}

\subsection{Legal \& Ethical Considerations} 
\begin{itemize}

\item \textit{If the dataset relates to people (e.g., their attributes) or was generated by people, were they informed about the data collection? (e.g., datasets that collect writing, photos, interactions, transactions, etc.)} N/A
    \item \textit{If it relates to other ethically protected subjects, have appropriate obligations been met? (e.g., medical data might include information collected from animals)} N/A
    \item \textit{If it relates to people, were there any ethical review applications/reviews/approvals? (e.g. Institutional Review Board applications)}N/A
    \item \textit{If it relates to people, were they told what the dataset would be used for and did they consent? What community norms exist for data collected from human communications? If consent was obtained, how? Were the people provided with any mechanism to revoke their consent in the future or for certain uses?} N/A
    \item \textit{If it relates to people, could this dataset expose people to harm or legal action? (e.g., financial social or otherwise) What was done to mitigate or reduce the potential for harm?} N/A
    \item \textit{If it relates to people, does it unfairly advantage or disadvantage a particular social group? In what ways? How was this mitigated?} N/A
    \item \textit{If it relates to people, were they provided with privacy guarantees? If so, what guarantees and how are these ensured?} 
N/A

\item \textit{Does the dataset comply with the EU General Data Protection Regulation (GDPR)? Does it comply with any other standards, such as the US Equal Employment Opportunity Act?} 
N/A

\item \textit{Does the dataset contain information that might be considered sensitive or confidential? (e.g., personally identifying information)} 
No.

\item \textit{Does the dataset contain information that might be considered inappropriate or offensive?} 
No.
\end{itemize}

\end{document}